\documentclass{article} 
\usepackage{preprint,times}

\usepackage{amsmath,amsfonts,bm}

\def\eqref#1{equation~\ref{#1}}
\def\Eqref#1{Equation~\ref{#1}}

\def\1{\bm{1}}

\DeclareMathAlphabet{\mathsfit}{\encodingdefault}{\sfdefault}{m}{sl}
\SetMathAlphabet{\mathsfit}{bold}{\encodingdefault}{\sfdefault}{bx}{n}

\usepackage{amssymb,amsthm}
\usepackage{graphicx}
\usepackage{booktabs}

\theoremstyle{plain}
\newtheorem{theorem}{Theorem}
\newcommand{\Amb}{\operatorname{Amb}}

\usepackage[hidelinks]{hyperref}
\usepackage{url}

\newif\ifpreprintversion
\newcommand{\preprintcopy}{\preprintfinaltrue\preprintversiontrue}

\title{AmbiModBench: Benchmarking Gene Perturbation Prediction Beyond Shared Responses}

\preprintcopy   

\author{
Sikai Huang$^{1}$\thanks{Equal contribution.}
\hspace{0.12in}
Zhiwen Yang$^{2,\ast}$\thanks{Corresponding authors: Zhiwen Yang
(\texttt{zhiwen.yang@connect.polyu.hk}) and Stan Z.\ Li
(\texttt{Stan.ZQ.Li@westlake.edu.cn}).}
\hspace{0.12in}
Kai Yu$^{3}$
\hspace{0.12in}
Jiayuan Chen$^{1}$
\hspace{0.12in}
Stan Z. Li$^{3,\dagger}$
\\[0.15cm]
$^{1}$Fujian Agriculture and Forestry University\\[0.02cm]
$^{2}$The Hong Kong Polytechnic University\\[0.02cm]
$^{3}$Westlake University
}

\begin{document}

\maketitle

\ifpreprintversion\lhead{Preprint}\fi

\begin{abstract}
Predicting cellular responses to genetic perturbations helps prioritize experiments in single-cell genomics, where exhaustive measurement is infeasible. While computational models increasingly predict these responses, three evaluation deficiencies obscure what their scores demonstrate. First, absolute metrics cannot separate target-specific predictions from a shared background response. Second, common metrics remain high under gene shuffling, so gene-level accuracy is never verified. Third, a score at one training size says nothing about coverage, which depends on representation-space proximity and response-constraining power. We propose AmbiModBench, a specificity-aware, gene-resolved and coverage-aware benchmark. It pairs every score with a training-mean reference fitted on the same split, screens each readout by gene-coordinate permutation, and links embedding distance to response variation. Across K562, RPE1 and Norman, strong absolute scores largely reflect shared background rather than target-specific learning. Widely used readouts track response magnitude distributions rather than the affected genes. Detectable gain follows representation-space coverage rather than training-set size. Nonetheless, on RPE1 the protocol yields a reproducible target-specific gain across five additional splits and three gene selections, which absolute scores alone cannot distinguish from shared background.

\end{abstract}

\section{Introduction}
\label{sec:intro}

Perturbing cells with genetic interventions and reading out the
transcriptome reveals how genes shape cellular state, and the resulting
maps are used to nominate disease drivers and therapeutic targets
\citep{replogle2022mapping,norman2019exploring}. CRISPR screening and
laboratory automation have moved these experiments to genome scale, so
that tens of thousands of knockdowns can be measured in a single
experiment \citep{replogle2022mapping}. Even at that scale, the space of
possible interventions far exceeds what any screen can assay:
protein-coding genes taken individually and in combination, across cell
types and cellular contexts, cannot be measured exhaustively
\citep{wu2025perturbench}. Predicting the response to an unmeasured
perturbation is therefore not only a modelling problem, but a practical
way to decide which experiment to run next.

Models now use latent cellular states, gene relationships and
protein-sequence representations to predict perturbation responses
\citep{lotfollahi2019scgen,lotfollahi2023cpa,roohani2023gears,lin2023esm,su2025protrek},
and the 2026 Virtual Cell Challenge now asks for zero-shot prediction in
cellular contexts that provide no perturbation measurements at all
\citep{arc2026vcchallenge}. Reported gains remain difficult to compare:
datasets, response definitions, held-out units and metric suites differ
across studies, so a benchmark serves not merely to rank models but to
identify which predictions carry information that is useful for
selecting unmeasured interventions
\citep{wei2026benchmarking,li2026systematic}. Recent evaluations have
established the importance of simple baselines and systematic variation
\citep{ahlmanneltze2025dl,csendes2025foundation,vinas2025systema}.
Despite this progress, \textbf{three deficiencies} still hinder the
further development of perturbation response prediction:

\begin{itemize}
\setlength{\itemsep}{2pt}
\setlength{\parsep}{0pt}
\setlength{\parskip}{0pt}
\item \textbf{Absolute scores without a reference.} An absolute score
cannot distinguish shared-response recovery from target-dependent
prediction. Different perturbations induce a substantial common transcriptional
response, so a model can capture this component and score highly
without distinguishing the intervention being evaluated.
Baseline equivalence of this kind has been reported for foundation
models \citep{ahlmanneltze2025dl,csendes2025foundation}, and systematic
variation inflates conventional scores
\citep{vinas2025systema}.
An absolute score does not reveal how much it exceeds a reference
obtained by averaging training perturbations.

\item \textbf{Metrics blind to gene assignment.} Metrics for
gene-specific prediction are often not validated for
gene identity. A response vector associates each predicted change with a
particular gene, and a score computed only from the distribution of its
values discards that association: shuffling the values among gene labels leaves
the score unchanged.
Energy and Wasserstein distances over marginal response values have
this invariance: they can measure agreement in response magnitudes
without establishing that the affected genes were predicted correctly.
Metric calibration through positive and negative controls has been
proposed to check what a score rewards before models are compared
\citep{miller2025calibration}. The screen here tests an exact invariance
to gene assignment rather than a calibration range, and pairs it with
calibration of the response to measurement noise.
Gene-specific prediction requires both.

\item \textbf{Coverage reduced to sample size.} Training-set size alone
is not a response-relevant coverage condition. For a held-out perturbation, the available evidence depends on the
proximity of the observed targets in representation space and on how
tightly nearby representations constrain response variation, not on
their number alone.
Two training sets of the same size can leave different gaps around
the same test target, and closely spaced embeddings can correspond to
vastly different responses when the representation misses the
biological mechanisms that determine them.
Context coverage, rather than model capacity, has been argued to be the
binding constraint on perturbation models
\citep{dibaeinia2026context}; the present formulation makes it
measurable at the representation level.
Reporting a score at one sample size leaves the geometric coverage and
the response-constraining power of the representation implicit, giving
little guidance about where predictions are supported.
\end{itemize}

\begin{figure}[t]
\centering
\includegraphics[width=\textwidth]{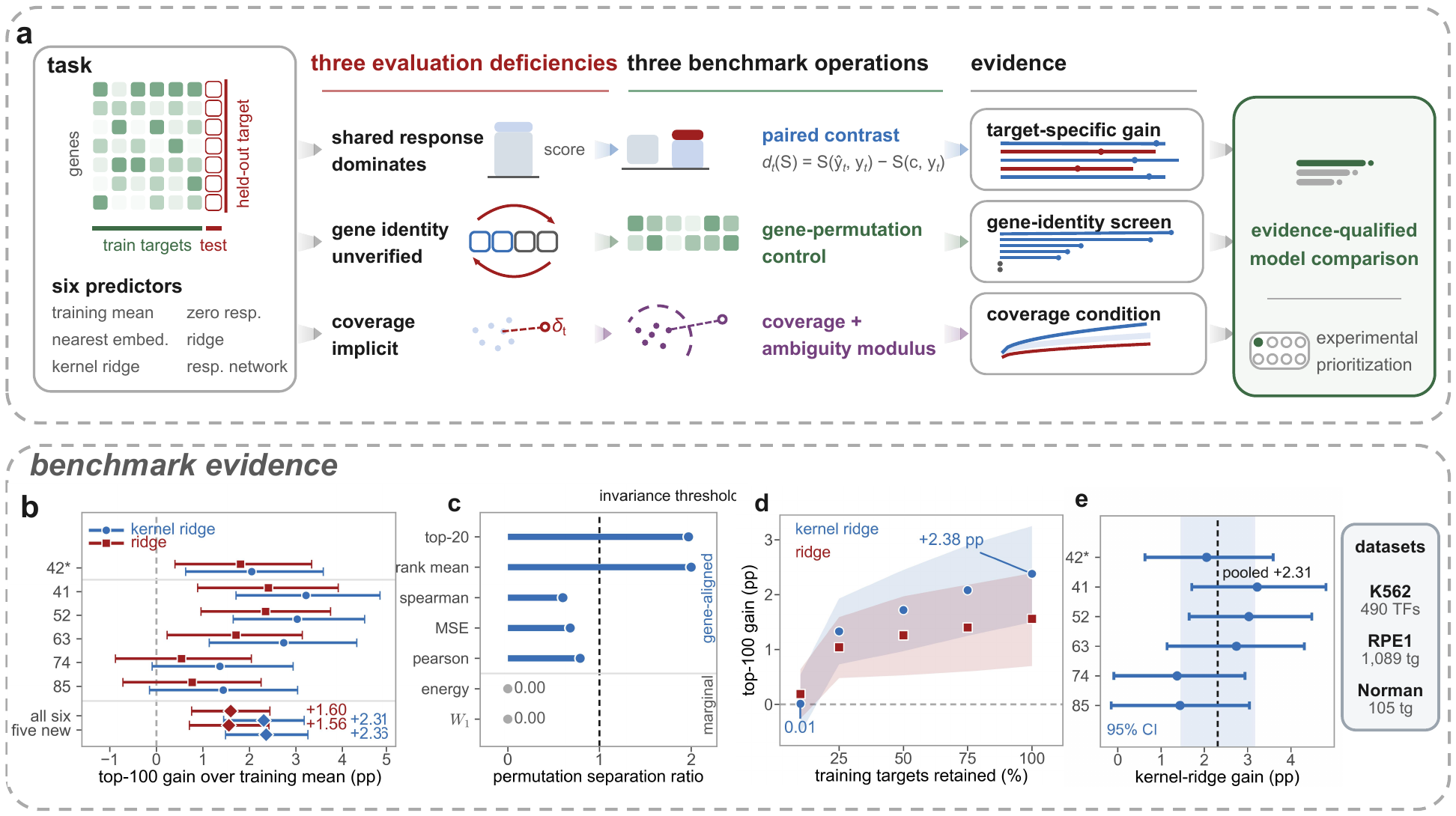}
\caption{AmbiModBench overview. (a) Benchmark design: three deficiencies,
the corresponding operations and the evidence they return. The
paired contrast scores a prediction and the training-mean reference
$c$ against the same held-out response $y_t$; the gene-permutation
control permutes gene coordinates; the coverage operation pairs the
test-to-training radius $\delta_t$ with the ambiguity modulus. (b)
Top-100 direction-agreement gains over the training mean for kernel
ridge and ridge across six RPE1 splits. (c) Gene-identity screen:
permutation-induced score separation relative to the reference-to-mean
gap, for seven K562 readouts; energy distance and Wasserstein distance
$W_1$ remain at zero. (d) The same paired gain as the retained
training fraction grows from 10\% to 100\%. (e) Kernel-ridge effects
from (b) with 95\% target-cluster bootstrap intervals. All evaluations
use target-disjoint holdouts.}
\label{fig:overview}
\end{figure}

To address these deficiencies, we introduce \textbf{AmbiModBench}
(Ambiguity-Modulus Benchmark; Figure~\ref{fig:overview}), an
evidence-qualified benchmark for evaluating gene-specific perturbation
prediction beyond the training mean. The ambiguity modulus bounds how
much responses can vary when representations are close and thereby
connects representation geometry with response variation.
Rather than serving only as another benchmark table, AmbiModBench is
designed as a paired protocol that brings a shared-response reference,
gene-assignment calibration and a coverage condition into one
comparable record.

\textbf{Contributions.} In this work, we address the three
deficiencies in the same order and \textbf{(1)} replace the absolute
score with a paired training-mean contrast, in which the reference is
fitted on training targets and scored against the same held-out
response, reported with target-cluster uncertainty; \textbf{(2)}
calibrate readouts by gene-coordinate permutation and independent cell
replicates, separating gene assignment from amplitude; and \textbf{(3)}
formulate training coverage through a test-to-training embedding radius
and an ambiguity modulus, with a conditional bound and a fixed-test
subsampling experiment.
Together, these components turn a benchmark result into an auditable
claim about target-specific prediction, gene-specific measurement and
the training evidence available for that claim.
We evaluate AmbiModBench on K562, RPE1 and Norman with six fixed
predictors under target-disjoint holdouts. Across six RPE1 splits, kernel
ridge exceeds the training-mean reference by 2.31 percentage points
[1.45, 3.18], and the gain grows with the observed training set. K562
sits at the reference, and Norman is unresolved because its interval
includes zero.

\section{Related Work}
\label{sec:related}

\textbf{Perturbation prediction models.} Perturbation prediction methods
encode different forms of biological
structure. scGen represents perturbation effects as shifts in a learned
latent space \citep{lotfollahi2019scgen}. CPA composes representations of
perturbations, cellular contexts and covariates
\citep{lotfollahi2023cpa}. GEARS uses gene relationships to predict
transcriptional responses to previously unobserved perturbations and
combinations \citep{roohani2023gears}. Protein language models offer a
complementary source of target information through sequence-derived
representations \citep{lin2023esm,su2025protrek}. These model families
differ in their native inputs and generalization settings, which makes
the evaluation reference and held-out unit central to comparison.

Zero-shot learning addresses the same prediction problem for unseen
categories using fixed representations
\citep{palatucci2009zero,lampert2014attribute}; AmbiModBench examines
this setting for perturbation targets.

\textbf{Datasets and benchmarks.} Large single-cell perturbation datasets
make held-out intervention
prediction measurable. The Replogle Perturb-seq atlas includes
genome-scale CRISPR interference in K562 and an essential-gene screen
in RPE1 \citep{replogle2022mapping}. The Norman CRISPR activation
screen measures single and combinatorial perturbations and their
transcriptional phenotypes \citep{norman2019exploring}. These resources
provide complementary coverage of target populations and intervention
types, and a benchmark spanning them must specify the response
definition and held-out task alongside the score.

Benchmarking has moved from comparing models with each other to
auditing what a score rewards. Simple baselines match published methods
\citep{ahlmanneltze2025dl,csendes2025foundation}, control-population
reuse inflates conventional scores \citep{nicol2026spurious}, and
metric calibration by positive and negative controls has been proposed
as a prerequisite for comparison \citep{miller2025calibration}. Systema
attributes this equivalence to systematic variation between perturbed
and control populations, and reweights evaluation toward
perturbation-specific effects \citep{vinas2025systema}. Concurrent
benchmarks vary predictors, datasets and metric suites: comparisons of
27 methods over 29 datasets and of 13 methods over 25 datasets agree
that different metrics probe distinct biological signals
\citep{wei2026benchmarking,li2026systematic}, rank metrics over a pool
of reference perturbations expose a mode collapse that fit-based
measures miss \citep{wu2025perturbench}, and gene-level signatures can
be lost despite favorable quantitative scores \citep{radig2026scarchon};
the 2026 Virtual Cell Challenge scores submissions with a composite of
Cell-Eval metrics \citep{arc2026vcchallenge}. These studies
establish that a score has to be interpreted rather than merely
maximized. None of them, however, compares a prediction against a
reference fitted on the same split, checks which metrics detect a wrong
gene assignment, or states whether a held-out target is covered at the
representation level. Sections~\ref{sec:ambimod}--\ref{sec:experiments}
supply and evaluate these three elements.

\section{Benchmark Design}
\label{sec:ambimod}
\label{sec:design}

\subsection{Task Definition}
\label{sec:setup}

Consider a fixed cellular context and perturbation targets
$t\in\mathcal{T}$. Let $\mu_t\in\mathbb{R}^G$ denote the population mean
response over $G$ genes and let $y_t=\mu_t+\epsilon_t$ be its observed
estimate. A target representation $z_t=\phi(t)\in\mathbb{R}^d$ is
available for training and held-out targets. Given responses for
$\mathcal{T}_{\mathrm{tr}}$, the task is to predict the response to
$t\in\mathcal{T}_{\mathrm{te}}$, where
$\mathcal{T}_{\mathrm{tr}}\cap\mathcal{T}_{\mathrm{te}}=\varnothing$.
The predicted response $\hat y_t\in\mathbb{R}^G$ is evaluated against
$y_t$. The held-out target is the unit of generalization, with cellular
context fixed within each dataset. Its response is excluded from model
fitting, representation transformation and hyperparameter selection.

\subsection{Benchmark Datasets}

Table~\ref{tab:design} summarizes the three experimental regimes.
\textbf{K562} contains 490 transcription factors from the Replogle atlas
\citep{replogle2022mapping}, with random, family and embedding-ball
holdouts over five seeds giving 15 evaluation instances.
\textbf{RPE1} contributes 1,089 eligible perturbation
targets with protein-sequence representations and finite responses over
4,063 common genes, split into 871 training and 218 evaluation targets;
the benchmark includes the locked seed42 split and five additional
splits. \textbf{Norman} contributes 105 eligible single-target
perturbations with 84/21 train/test splits under three seeds and 5,045
genes. The Splits column of Table~\ref{tab:design} counts evaluation
instances: 15 K562 holdout instances, six RPE1 target splits and three
Norman target splits, with K562 instance sizes depending on the holdout
design. Appendix~\ref{app:protocol} specifies response construction,
control sharing, target eligibility and gene alignment.

\begin{table}[t]
\centering
\small
\caption{Benchmark datasets and evaluation regimes.}
\label{tab:design}
\begin{tabular}{lrrrl}
\toprule
\textbf{Dataset} & \textbf{Targets} & \textbf{Genes} & \textbf{Splits} & \textbf{Response scale} \\
\midrule
K562 & 490 & 1,000 & 15 & log1p mean-count difference \\
RPE1 & 1,089 & 4,063 & 6 & mean-count difference \\
Norman & 105 & 5,045 & 3 & log-normalized difference \\
\bottomrule
\end{tabular}
\end{table}

\subsection{Predictors and Representations}

Every comparison is paired against a common reference, so the predictor
family and the target representation are fixed before the
additional-split evaluation. The representation also defines the
coverage condition of Section~\ref{sec:coverage_theory}, which is why
both are specified here.
The benchmark fixes six predictors: training mean, zero response,
nearest embedding, ridge, radial-basis-function kernel ridge and a
spectrally normalized response network. The training mean supplies the
evaluation reference. The zero predictor defines an unchanged-response
control and nearest embedding predicts the response of the most
similar training target. Ridge and kernel ridge form the primary
learned baseline family, fixed before the additional-split evaluation.
The response network supplies a neural readout under the same target
holdouts. All six predictors are reported in the supporting tables.

\textbf{Target representations.} RPE1 uses the pretrained
protein-sequence encoder of ProTrek-650M
\citep{su2025protrek}; K562 and Norman use ESM2 representations
\citep{lin2023esm}. For regression readouts, principal components are
fitted on outer-training targets, followed by row normalization, and
hyperparameters are selected without access to held-out direction
scores. Appendix~\ref{app:models}
reports dimensions, search grids and dataset-specific implementations.

\section{Evaluation Protocol}
\label{sec:protocol}

\subsection{Paired Training-Mean Contrast}
\label{sec:contrast}

The benchmark reference is the
mean response over training targets,
$c=|\mathcal{T}_{\mathrm{tr}}|^{-1}
\sum_{u\in\mathcal{T}_{\mathrm{tr}}}y_u$.
For a score $S$ oriented so that larger is better, AmbiModBench measures the
paired effect
\begin{equation}
d_t(S)=S(\hat y_t,y_t)-S(c,y_t),
\qquad
\widehat{\Delta}_S=\frac{1}{N}\sum_{(t,s)\in\mathcal{I}}d_{t,s}(S),
\label{eq:contrast}
\end{equation}
where $s$ indexes a split, $\mathcal{I}$ is the set of target-split
records and $N=|\mathcal{I}|$. The reference is fitted separately in
each split, so that the comparator sees exactly the training targets
available to the evaluated model. A positive effect measures
improvement over a response
that does not use the evaluated target's identity. The subtraction is
performed on scores, without assuming an additive decomposition of
the response vector. This paired
quantity is the common reporting unit across datasets, models and
gene selections.

\subsection{Direction Agreement and Inference}
\label{sec:statistics}

Because an absolute score carries no reference, every reported quantity
is the paired effect of \Eqref{eq:contrast} against the
reference of Section~\ref{sec:contrast}, which also fixes the sampling
unit of the intervals below.
\textbf{Direction agreement.} This readout
measures whether a prediction assigns the observed
sign to the strongest response genes. Let $J_k(y_t)$ contain the $k$
largest entries of $|y_t|$, with deterministic tie breaking. Define
\begin{equation}
D_k(\hat y_t,y_t)=\frac{1}{k}\sum_{g\in J_k(y_t)}
\left[
\mathbf{1}\{\hat y_{tg}\ne0,\ 
\operatorname{sign}(\hat y_{tg})=\operatorname{sign}(y_{tg})\}
+\tfrac12\mathbf{1}\{\hat y_{tg}=0\}
\right].
\label{eq:direction}
\end{equation}
A zero prediction receives half credit. Positive rescaling leaves the
score unchanged. The gene set is determined only during scoring and is
never supplied to a predictor. It contains the strongest measured
responses, not significance-filtered differentially expressed genes.

\textbf{Gene selection.} RPE1 uses $D_{100}$ of
\Eqref{eq:direction} as its primary score. For cross-dataset comparisons,
we set $k=\max(20,\operatorname{round}(0.02G))$ to account for gene-space
size. This selects 20, 81 and 101 genes in K562, RPE1 and Norman.
Top 20, top 100 and top 2\% are reported for every predictor in the
supporting results. Pearson correlation, rank correlation, rank mean,
top-20 recall and squared error supply complementary readouts in
measurement calibration.

\textbf{Target-cluster inference.} Model and training-mean scores are
paired by target and split. Pooled
intervals use 10,000 target-cluster bootstrap draws and retain all
split records of each resampled target. RPE1 contributes 1,308 records
from 792 unique test targets. The five-additional-split analysis
contributes 1,090 records from 724 targets. Individual-split and
coverage intervals use 2,000 draws. Coverage repeats are averaged
within target and split before inference. These intervals quantify
target sampling variation conditional on the fitted splits.
Appendix~\ref{app:statistics} specifies the resampling procedure,
stored bootstrap-tail summaries, multiplicity families and dataset
contrasts. Reported intervals are per-estimate and are not treated as
family-wise significance tests.

\subsection{Gene-Identity Calibration}
\label{sec:calibration}

Model comparison presupposes that the score responds to gene assignment;
Section~\ref{sec:measurement_results} supplies that check for the
readout used in the comparisons above. \textbf{Gene-permutation screen.} This screen tests whether a metric retains gene
identity. The screen compares an independent-replicate response estimate
with a copy whose gene coordinates are permuted. A change in score
establishes sensitivity to gene assignment. A metric computed solely
over the marginal distribution of values across genes is invariant to
this permutation. The screen concerns gene-specific response prediction.

\textbf{Independent-replicate calibration.} Replication separately
calibrates magnitude-sensitive scores. Write
$\mu_t=m+s_t$ for shared response $m$ and target-dependent component
$s_t$, and let independent samples give $q_t=\mu_t+\eta_t$ and
$y_t=\mu_t+\epsilon_t$ with zero-mean independent errors. The mixture
$p_t(a)=m+a(q_t-m)$, $0\le a\le1$, has expected squared error
\begin{equation}
\mathbb{E}\|p_t(a)-y_t\|^2
=(1-a)^2\|s_t\|^2+a^2V_q+V_y,
\label{eq:shrinkage}
\end{equation}
with $V_q=\mathbb{E}\|\eta_t\|^2$, $V_y=\mathbb{E}\|\epsilon_t\|^2$ and
optimum $a^\star=\|s_t\|^2/(\|s_t\|^2+V_q)$ when the denominator is
positive. An interior minimum reflects uncertainty in the replicate
reference. Appendix~\ref{app:proofs} extends the derivation to an
estimated shared response, retaining the covariance terms induced by
shared observations or controls.

\subsection{Training Coverage}
\label{sec:coverage_theory}

\textbf{Ambiguity modulus and coverage radius.} The number of observed
targets does not state whether a held-out target
is supported, which is the deficiency this section addresses. Training
coverage concerns the location of observed targets and the
response information their representations preserve. A test target near
the training set is well supported only when nearby embeddings constrain
the range of possible responses. To separate these two conditions,
define the ambiguity modulus and test-to-training radius as
\begin{equation}
\Amb(\phi,\mu)=
\sup_{t\ne u}\frac{\|\mu_t-\mu_u\|_2}{\|\phi(t)-\phi(u)\|_2},
\qquad
\delta_t=\min_{u\in\mathcal{T}_{\mathrm{tr}}}\|z_t-z_u\|_2.
\label{eq:ambiguity}
\end{equation}
Identical embeddings with distinct mean responses give an infinite
ratio. Identical embeddings with identical responses contribute zero.
A finite bound $\Amb(\phi,\mu)\le L$ states response-specific
smoothness. The radius $\delta_t$ measures the distance from a test
target to the observed training set. A small radius represents local
training support, while a small modulus limits response differences
within that neighborhood. Neither condition is specified by the
number of training targets alone.

\begin{theorem}[Conditional coverage bound]
\label{thm:coverage}
Assume $\|\mu_t-\mu_u\|_2\le L\|z_t-z_u\|_2$. Let $u^\star$ be a nearest
training target chosen using embeddings alone. If its response estimate
is unbiased with error variance at most $v$, then
$\hat\mu_t=y_{u^\star}$ satisfies
\[
\mathbb{E}\|\hat\mu_t-\mu_t\|_2^2\le L^2\delta_t^2+v.
\]
For noiseless training responses equal to zero, every estimator has
worst-case squared error at least $L^2\delta_t^2$ over this Lipschitz
class.
\end{theorem}

The nearest-neighbor and two-point arguments are standard
\citep{tsybakov2009introduction}. Appendix~\ref{app:proofs} gives the
proof. The bound separates coverage from response-estimation noise
under the stated assumption. Smaller $\delta_t$ and $L$ tighten its
approximation term. The relevant quantity is their joint contribution,
not training-set size alone.

\textbf{Fixed-test subsampling.} This experiment measures how directional
gain depends on the
observed training set. Each of the five additional RPE1 splits uses
five random orderings of its 871 training targets, and nested prefixes
retain 10\%, 25\%, 50\%, 75\% and 100\% of each ordering.
The training mean, representation transformation and readouts are
refitted at every fraction, giving 25 split/repeat runs per fraction.
For a fixed representation, adding targets cannot increase the radius
$\delta_t$ of \Eqref{eq:ambiguity}, so the experiment evaluates the full refitted
procedure as these observations become available.

\section{Experiments}
\label{sec:experiments}

The experiments address the three deficiencies of
Section~\ref{sec:intro} in the same order: Sections~\ref{sec:rpe1}
and~\ref{sec:measurement_results} test the paired contrast and the
gene-identity calibration, Section~\ref{sec:coverage_results} the
coverage condition, and Section~\ref{sec:crossdataset} places the
outcome in three regimes.

\subsection{High Absolute Scores Do Not Imply Target-Specific Prediction}
\label{sec:rpe1}

\textbf{High absolute scores do not imply target-specific prediction.}
RPE1 provides the positive test of the paired
evaluation.
Kernel ridge improves top-100 direction agreement over the training mean
by 2.31 percentage points, with a 95\% target-cluster bootstrap
interval of
[1.45, 3.18]. In these units the gain counts how many additional genes,
among the 100 strongest measured responses and averaged over targets,
receive the correct sign. These estimates average the paired effects of
\Eqref{eq:contrast} over held-out target-split records.
Table~\ref{tab:primary_full} reports the ridge contrast for every gene
selection, and Table~\ref{tab:all_rpe1} adds the remaining predictors:
ridge separates from the reference by a smaller margin, while the
zero-response and nearest-embedding baselines have negative point
estimates.

\begin{figure}[t]
\centering
\includegraphics[width=\textwidth]{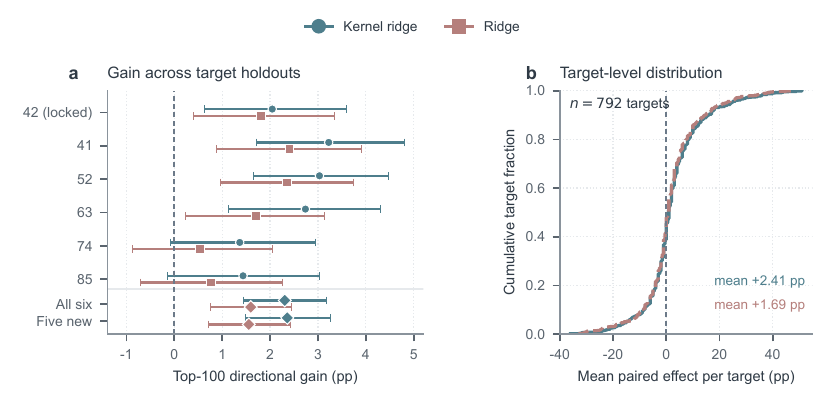}
\caption{RPE1 directional gains reproduce across held-out splits.
Effects are in percentage points (pp) relative to the training mean;
positive values indicate a gain.
(a) Points show paired top-100 effects for each split, with 95\%
target-cluster bootstrap
intervals. Diamonds show pooled effects for all six splits and for the
five additional splits excluding seed42. Each split has 218 test
targets. Pooled intervals cluster by target across repeated appearances.
(b) Empirical cumulative distributions of target-specific effects,
averaged over available test appearances for each of 792 targets.
These descriptive distributions weight each unique target equally,
while panel (a) pools target-split records.}
\label{fig:replication}
\end{figure}

\textbf{The gain is not tied to the locked split.}
After excluding seed42, kernel ridge yields 2.36 [1.49, 3.26] percentage
points, every additional split has a positive point estimate, and ridge
behaves the same way (Figure~\ref{fig:replication}a). The gain is
therefore not an artefact of the split that was previously locked.

\textbf{The gain is not tied to one gene cutoff.} Across all six splits,
kernel-ridge gains are 2.45 percentage points for top 2\% and 3.57 for
top 20, ridge follows the same ordering, and all four intervals exclude
zero (Appendix~\ref{app:fullresults}).
The coordinated comparison of Section~\ref{sec:crossdataset}
places the same paired effect in two further regimes, where the K562
estimates sit near the training mean (Table~\ref{tab:all_k562}).

\subsection{Magnitude-Based Scores Do Not Verify Gene Assignment}
\label{sec:measurement_results}

\textbf{Not every score detects a wrong gene assignment.} K562 provides
disjoint cell partitions, a noisy response reference and a separate
evaluation response per target.

\begin{figure}[t]
\centering
\includegraphics[width=\textwidth]{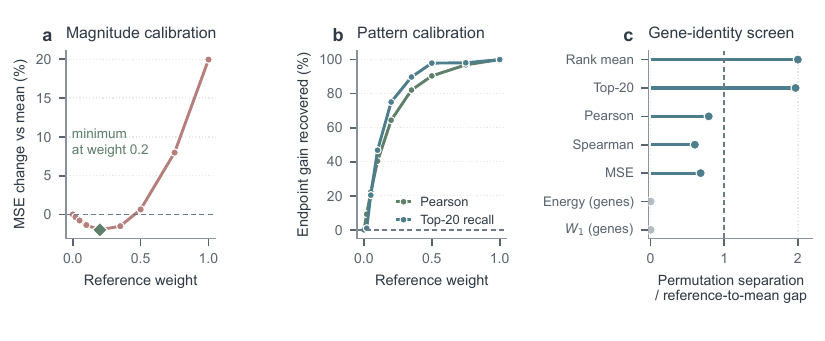}
\caption{Calibration identifies which response information a score uses.
(a) K562 squared error along a mixture of the prediction-side cohort mean and an
independent-replicate reference, with its minimum at weight 0.2.
(b) Pearson correlation and top-20 recall increase along the same
mixture. Each curve is normalized by its own endpoint gain.
(c) Score separation after permuting gene coordinates, divided by the
score gap between the replicate reference and the cohort mean. Zero
denotes invariance to
gene assignment. Calibration uses all 490 K562 TFs, prediction
replicates 0--2, evaluation replicates 3--4 and five gene permutations
per target, reused for both evaluation replicates.
Curves show descriptive cohort averages. The cohort mean is used for
measurement calibration, separately from the held-out-target evaluation.}
\label{fig:calibration}
\end{figure}

\textbf{Magnitude and pattern scores disagree on one prediction.}
In the independent-replicate experiment, squared error first decreases
and then increases as the reference weight grows
(Figure~\ref{fig:calibration}a).
The minimum occurs at $a=0.2$, as expected when the reference estimate
contains noise. At the same time, Pearson correlation and top-20 recall
increase across the mixture (Figure~\ref{fig:calibration}b).
That minimum is the mechanism behind Section~\ref{sec:rpe1}: squared
error falls when a prediction is shrunk toward a noisy cohort mean, so a
model can improve this score without placing the response on the correct
genes, which is exactly the failure an absolute score hides. This is
consistent with \Eqref{eq:shrinkage}: the scores value different
properties of a finite-sample response. It supports reporting a
scale-invariant readout alongside squared error
when the scientific question concerns response pattern.

\textbf{Marginal-distribution distances ignore gene identity.}
Five gene-aligned readouts, namely top-20 recall, rank mean, rank
correlation, Pearson correlation and squared error, all respond to
scrambling gene assignments.
Energy distance and Wasserstein distance over marginal values, the
empirical distribution of response magnitudes that discards gene
identity, remain exactly unchanged (Figure~\ref{fig:calibration}c).
The separation is largest for the two rank-based readouts, smaller for
the correlation and error measures, and exactly zero for the two
marginal distances.
A benchmark that reports only such distances cannot testify to
gene-level correctness at any score level, because the quantity it
evaluates is unchanged by the assignment it would need to test.
The screen therefore makes gene assignment a criterion for choosing a
readout rather than an assumption; the readout in
Section~\ref{sec:rpe1} retains aligned gene coordinates.
The response network, the only neural predictor in the suite, does not
separate from the reference in any of the three datasets, with a paired
interval at the primary selection narrower than the change produced by a
single additional correct gene in one held-out record
(Tables~\ref{tab:all_rpe1}--\ref{tab:all_norman}).
The score depends only on the sign pattern of the selected genes, so a
prediction proportional to the reference is not separated from it.
Its residual amplitude is set by its normalization
(Appendix~\ref{app:ablation}).

\subsection{Detectable Gain Depends on Coverage, Not Training-Set Size}
\label{sec:coverage_results}

\begin{figure}[t]
\centering
\includegraphics[width=0.93\textwidth]{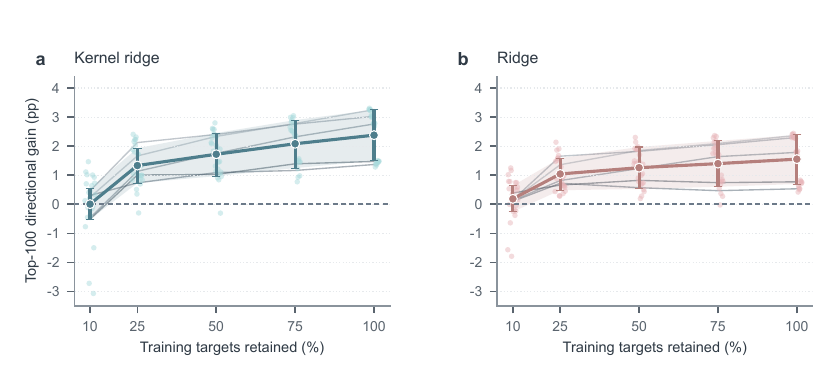}
\caption{Increasing the observed training set strengthens the RPE1
directional signal. Gains are in percentage points (pp) relative to
the refitted training mean. Test targets are fixed within each of five splits.
Pale points show all five subset-ordering repeats per split at each
fraction. Gray trajectories average the repeats within each split.
Colored points and ribbons show pooled paired effects and 95\%
target-cluster bootstrap intervals after averaging repeats within target and
split. Kernel ridge (a) and ridge (b) use the same nested training
subsets, with all transformations and references refitted.}
\label{fig:coverage}
\end{figure}

\textbf{Equal training sizes support different gains.}
Kernel-ridge gain rises from 0.01 percentage
points at 10\% of the training set to 1.33, 1.72, 2.08 and 2.38 at the
successive fractions, and ridge follows the same ordering
(Figure~\ref{fig:coverage}, Table~\ref{tab:coverage_full}).
At 10\% of the training targets the paired effect is 0.01 percentage
points with an interval that contains zero: at this coverage the same
model is indistinguishable from the training mean, and the gain becomes
detectable only between 10\% and 25\%.

\textbf{Detectable gain grows with the observed training set.}
The held-out task and scoring rule stay fixed, so the trajectory makes
training support part of the benchmark result. Training fraction serves
here as an operational proxy for training coverage. For
a fixed representation and test target, adding training targets cannot
increase the nearest training distance (Section~\ref{sec:coverage_theory}).
Therefore, larger fractions move the evaluation along a coverage axis
rather than merely a size axis.
The curve therefore specifies how much additional response data changes
the paired gain within this regime, which is the quantity a user needs
when deciding how many perturbations to observe.

\subsection{Cross-Dataset Evaluation Regimes}
\label{sec:crossdataset}

\textbf{The same protocol returns different verdicts in different regimes.}
Applying one score formula and one primary method family to all three
datasets separates a regime with a positive paired gain from one whose
readouts sit at the training mean, while the CRISPRa regime remains
unresolved because its interval includes zero.
Absolute scores should therefore be read against the measurement and
coverage conditions of each regime rather than averaged into a single
verdict. Appendix~\ref{app:regimes} reports the coordinated comparison,
including the per-dataset gains and a direct RPE1-minus-K562 contrast
that excludes zero for both primary methods.

\section{Discussion and Conclusion}
\label{sec:conclusion}

\textbf{Summary.} AmbiModBench makes perturbation-prediction scores
informative only after
three conditions are explicit: the gain beyond a shared-response
reference, the gene-specific information retained by the score, and the
training coverage supporting the held-out target. The paired
training-mean contrast addresses the first, gene-permutation and
replicate calibration the second, and the coverage radius, ambiguity
modulus and fixed-test subsampling the third. The coverage condition is
not specific to perturbation screens: it applies to any held-out
prediction task whose test items are described by a representation,
which is why we state it as a bound.

\textbf{Discussion.} The RPE1 result provides a concrete positive
outcome for this evaluation
logic: simple sequence-embedding readouts yield a positive directional
gain across repeated target holdouts, and the gain increases as more
training perturbations are observed. This identifies an experimental
setting in which additional response data supports detectable
out-of-target prediction. The same score formula returns no detectable
advantage over the training mean in K562, so a reported gain has to be
read together with the regime in which it was measured.
The resulting benchmark record includes the comparator, the measured
biological information, uncertainty and the training-coverage condition.
Together, these elements provide a closed, auditable basis for choosing
models and training conditions for further experimental testing.

\textbf{Limitations.} AmbiModBench qualifies a score rather than
maximizing one, and this scope has costs. The coverage bound of
Theorem~\ref{thm:coverage} assumes a finite Lipschitz constant that we
do not estimate, so it stays qualitative; estimating the ambiguity
modulus and the radius $\delta_t$ empirically remains open. The
benchmark fixes one family of sequence-derived representations, and the
coordinated comparison spans three datasets whose Norman interval
includes zero, reported as unresolved rather than negative evidence.
One predictor, the response network, is trained under a single fixed
configuration rather than a searched grid; removing its auxiliary
response-to-embedding penalty restores a response-aligned component in
the residual but not an effect the size of a single additional correct
gene, so the reported verdict is unchanged.
The paired contrast is reported for the direction readout; extending it
to effect-size and differential-expression scores is left to future
work.

\section*{AI Use Statement}

In this work, we used generative AI tools for the following tasks that
require disclosure: providing feedback on the design of the evaluation
protocol; formulating and revising the mathematical arguments, whose
complete proofs appear in Appendix~\ref{app:proofs}; and interpreting
returned experiment summaries. We have not used generative AI tools to
generate synthetic data sets---quantitative plots and tables are produced
from experimental output files---to clean or reformat the analyzed data
sets, to assist with translation, or to support qualitative or thematic
data analysis. The remaining required-disclosure tasks did not involve
generative AI tools in this work.

Additionally, we used generative AI tools for these recommended-disclosure
tasks: restructuring and editing the manuscript; literature and reference
checks; preparing analysis, visualization and document-validation
scripts; generating parts of the static benchmark-overview figure from a
protocol specification; and inspecting scientific-figure references and
checking the rendered figure.

We have reviewed all AI-assisted work: the mathematical arguments were
checked against the complete proofs in Appendix~\ref{app:proofs}, the
literature and reference checks against the original sources, and the
figure labels and reported effects against the benchmark records; an
author adjusted the generated figure by hand in PowerPoint. We take
responsibility for the final content of this work, including text, claims
and artifacts produced with the aid of generative AI.

\section*{Reproducibility Statement}

Section~\ref{sec:design} defines the datasets and target holdouts, and
Section~\ref{sec:protocol} the primary scores.
Appendix~\ref{app:protocol} describes response
construction, representations, tuning, training subsampling and
target-cluster inference. Appendix~\ref{app:proofs} states the assumptions
and gives complete proofs of the mathematical claims.
The supporting tables report all six predictors under the three
direction-score selections. The accompanying analysis records include
the locked configuration, per-target scores and hyperparameter choices
used to generate these summaries. Code and data are available at
\url{https://anonymous.4open.science/r/ambimod-4D5B/}.

\section*{Ethics Statement}

This study analyzes existing genetic-perturbation datasets and does not
collect new human-subject data. The reported results concern cellular
response prediction under the specified laboratory contexts. They do
not establish clinical efficacy or safety of interventions.

\clearpage
\bibliographystyle{preprint}
\bibliography{refs}

\clearpage
\appendix
\section{Coverage and Measurement Derivations}
\label{app:proofs}

\subsection{Proof of the conditional coverage bound}

Fix the training embeddings and a test embedding $z_t$. Let $u^\star$
minimize $\|z_t-z_u\|_2$ over training targets, with ties resolved without
using responses. Write $y_{u^\star}=\mu_{u^\star}+\epsilon_{u^\star}$.
Under the conditional zero-mean error assumption, expansion gives
\begin{align}
\mathbb{E}\|y_{u^\star}-\mu_t\|_2^2
&=\|\mu_{u^\star}-\mu_t\|_2^2
  +\mathbb{E}\|\epsilon_{u^\star}\|_2^2\\
&\leq L^2\|z_{u^\star}-z_t\|_2^2+v
 =L^2\delta_t^2+v.
\end{align}
The cross term vanishes by unbiasedness. The bound concerns prediction
of the population mean. Scoring against an independent noisy response
adds that response's error variance.

For the lower bound, let $Z_{\mathrm{tr}}$ denote the finite set of
training embeddings and choose a unit vector $e\in\mathbb{R}^G$.
The distance function $h(z)=\operatorname{dist}(z,Z_{\mathrm{tr}})$ is
1-Lipschitz by the triangle inequality. The two functions
\begin{equation}
f_+(z)=Lh(z)e,\qquad f_-(z)=-Lh(z)e
\end{equation}
are consequently $L$-Lipschitz, are zero at every training embedding,
and take values $L\delta_t e$ and $-L\delta_t e$ at the test embedding.
An estimator sees identical noiseless training data under these two
functions. For its output $a\in\mathbb{R}^G$,
\begin{equation}
\frac{\|a-L\delta_t e\|_2^2+\|a+L\delta_t e\|_2^2}{2}
=\|a\|_2^2+L^2\delta_t^2\geq L^2\delta_t^2.
\end{equation}
At least one squared loss is no smaller than this average. Taking an
expectation over a randomized estimator preserves the inequality.
Taking the infimum over estimators proves the stated minimax lower
bound for the zero-response training set.

Theorem~\ref{thm:coverage} is conditional on a fixed representation and a
finite response-specific Lipschitz constant. It does not estimate that
constant from noisy pairwise ratios. Its purpose is to make the
coverage scale explicit. The regression experiments evaluate fitted
readouts separately from this nearest-response construction.

\subsection{Shrinkage toward a known shared response}

Suppress the target subscript and write $\mu=m+s$, $q=\mu+\eta$ and
$y=\mu+\epsilon$. For $p(a)=m+a(q-m)$,
\begin{equation}
p(a)-y=-(1-a)s+a\eta-\epsilon.
\end{equation}
When $\eta$ and $\epsilon$ are independent, zero-mean errors,
the expected cross terms vanish. This gives
\Eqref{eq:shrinkage}. Differentiating with respect to $a$ gives
$-2(1-a)\|s\|_2^2+2aV_q$, so the minimizer on $[0,1]$ is
$a^\star=\|s\|_2^2/(\|s\|_2^2+V_q)$ when the denominator is positive.
If both terms are zero, every value of $a$ has the same risk.
Dividing all squared norms and variances by $G$ gives the corresponding
per-gene mean squared error.

\subsection{Estimated shared responses and shared controls}

An empirical shared response is itself estimated. Let
$\hat m=m+\xi$, where $\xi$ has mean zero and
$V_c=\mathbb{E}\|\xi\|_2^2$. For
$\hat p(a)=(1-a)\hat m+aq$, the error becomes
$-(1-a)s+(1-a)\xi+a\eta-\epsilon$.
Define the error cross-moments
$C_{cq}=\mathbb{E}\langle\xi,\eta\rangle$,
$C_{cy}=\mathbb{E}\langle\xi,\epsilon\rangle$ and
$C_{qy}=\mathbb{E}\langle\eta,\epsilon\rangle$.
Expanding the squared error yields
\begin{align}
\mathbb{E}\|\hat p(a)-y\|_2^2
={}&(1-a)^2\|s\|_2^2+(1-a)^2V_c+a^2V_q+V_y\nonumber\\
&+2a(1-a)C_{cq}-2(1-a)C_{cy}-2aC_{qy}.
\label{eq:estimated_mean}
\end{align}
The prediction-side cohort mean and replicate reference can share
observations, and response estimates can share control cells.
These cross-moments must be retained unless the sampling design removes
the corresponding dependence. Figure~\ref{fig:calibration} reports the
observed calibration curve. It does not infer latent variances or a
population-optimal mixture weight from the simplified model.

\subsection{Gene-permutation invariance}

Associate a response vector $x\in\mathbb{R}^G$ with its empirical
distribution of coordinate values,
$\nu_x=G^{-1}\sum_{g=1}^G\delta_{x_g}$, where $\delta_b$ is a unit point
mass at $b$. For any coordinate permutation $\pi$,
$\nu_{\pi x}=\nu_x$. Any distance computed solely between $\nu_x$ and
$\nu_y$ is consequently unchanged if either response's gene coordinates
are permuted. This includes the one-dimensional Wasserstein and energy
distances used in the screen. Distances on joint distributions of cell
vectors retain a different information structure and are not covered
by this invariance statement.

\section{Reproducible Evaluation Protocol}
\label{app:protocol}

\subsection{Response construction and target eligibility}

The K562 evaluation uses a 490-transcription-factor cohort from the
Replogle Perturb-seq atlas \citep{replogle2022mapping}. Perturbed cells
are divided into five disjoint partitions per target, retaining up to
100 cells in each partition and requiring at least 20. Control cells
are divided into five disjoint partitions of 1,000 cells. A control
partition is shared across targets within its replicate. For each gene,
the response is the difference between the log1p of the mean perturbed
count and the log1p of the mean control count. It is not the mean of
single-cell log differences. The 1,000 genes are selected by the
variance of log control pseudobulk means. Partitioning uses seed~20260730. Fitting averages the five responses of each training target.
The coordinated direction evaluation also scores against the
five-replicate mean. The calibration in
Appendix~\ref{app:k562} uses separate prediction and evaluation
partitions instead.

K562 uses seeds 41, 52, 63, 74 and 85 for random, DNA-binding-domain
family and embedding-ball holdouts. Random holdouts contain 98 targets.
The family holdout evaluates the largest annotated family containing at
least 30 members among the 490 targets. The embedding-ball holdout selects the 98 targets nearest
to a seed-selected embedding center. The archived split memberships
preserve the generator's random-state progression. The family membership
is fixed across seeds, and its repeated evaluations are clustered by
target in the pooled analysis.

RPE1 responses are mean stored counts in perturbed cells minus the
mean of all non-targeting control cells. Each eligible perturbation
has at least three cells, a finite response and an available protein
embedding. The resulting 1,089 targets are not restricted to
transcription factors. The gene space is the intersection of the
prediction-file headers and response columns, preserving response-column
order, with 4,063 genes. A shared control mean is used across targets.
Seeds 42, 41, 52, 63, 74 and 85 define six 871/218 target splits.
The locked seed42 split was previously evaluated. The
five-additional-split analysis excludes it.

The Norman analysis uses single-target CRISPRa perturbations from
\citet{norman2019exploring}. Labels of the form $t+\mathrm{ctrl}$ or
$\mathrm{ctrl}+t$ identify the single perturbation $t$. Responses are
differences between the mean supplied log-normalized expression and
the shared all-control mean. Eligibility requires at least ten
perturbed cells and an available embedding. The analysis retains
105 targets and 5,045 genes. Seeds 42, 43 and 44 define 84/21 splits,
contributing 49 unique evaluation targets. These response definitions
are retained when coordinating the scoring rule across datasets.

\subsection{Representations and readouts}
\label{app:models}

RPE1 uses 1,024-dimensional protein-sequence representations from the public
ProTrek-650M release \citep{su2025protrek}.
The protein encoder's projected representation
is extracted with \texttt{get\_protein\_repr}. K562 uses 1,280-dimensional
ESM2-650M layer-33 representations, averaged over amino-acid residues.
Norman uses the same 1,280-dimensional ESM2-650M layer-33
residue-averaged representations \citep{lin2023esm}.
No held-out perturbation response is used to construct these embeddings.

For the regression readouts, principal component analysis (PCA) is
fitted only on outer-training targets. K562 and RPE1 retain
$\min(128,n_{\mathrm{tr}}-1,d)$ components, where $n_{\mathrm{tr}}$ is
the training-target count and $d$ is the embedding dimension. Norman
retains 64 components. Transformed rows are normalized to unit length.
Kernel validation uses this outer-training transformation without
refitting PCA within each inner fold. The outer test responses remain
excluded throughout model selection.

Ridge uses an intercept and \texttt{RidgeCV} with 17 logarithmically
spaced regularization strengths from $10^{-4}$ to $10^4$, using its
default efficient leave-one-out criterion. Kernel ridge uses
$K(z,z')=\exp(-\gamma\|z-z'\|_2^2)$ and no explicit intercept.
Three shuffled validation folds with seed~17 select $(\alpha,\gamma)$
by mean validation squared error. Table~\ref{tab:grids} gives the grids.
All six full-training RPE1 splits select ridge strength
$\alpha=3.1623$ and kernel parameters $(\alpha,\gamma)=(1,0.5)$.

\begin{table}[htbp]
\centering
\caption{Kernel-ridge search grids. Hyperparameters are selected from
training targets, without access to held-out direction scores.}
\label{tab:grids}
\small
\begin{tabular}{lll}
\toprule
Dataset & Regularization $\alpha$ & Kernel coefficient $\gamma$ \\
\midrule
RPE1 & $\{0.001,0.1,1\}$ & $\{0.005,0.02,0.1,0.5\}$ \\
K562 & $\{0.001,0.1,1,10\}$ & $\{0.05,0.2,1,5\}$ \\
Norman & $\{0.001,0.1,1\}$ & $\{0.05,0.2,1\}$ \\
\bottomrule
\end{tabular}
\end{table}

The training mean averages training-target responses with equal target
weights. The zero predictor returns zero for every gene.
Nearest embedding returns the response of the most similar training
target. RPE1 selects this target by cosine similarity of the original
normalized ProTrek representations. K562 and Norman select it after
PCA and normalization.

The response network has four spectrally normalized linear layers
with hidden width 256 and ReLU activations between layers. Context and
dose inputs are fixed at zero and one. AdamW uses learning rate
$10^{-3}$ and weight decay $10^{-4}$. Training runs for at most 400
epochs, with validation every ten epochs and patience eight. The inner
validation set has $\max(30,\operatorname{round}(0.15n_{\mathrm{tr}}))$
targets. Responses are standardized using inner-training targets.
The selected epoch count is used to refit the network on the full
outer-training set. Its loss combines squared error, a spectral penalty
with coefficient 0.01 and a response-to-embedding distance-ratio penalty
with coefficient 0.05. The monotonicity penalty coefficient is zero.
The network seed is 42 for RPE1 and the split seed for K562 and Norman.
The archived implementation defines these auxiliary penalties exactly.
The primary inferential family comprises ridge and kernel ridge.

\subsection{Fixed-test training subsampling}

For each of the five additional RPE1 splits and each repeat
$r\in\{0,1,2,3,4\}$, the subset generator uses seed
$1000\,s+r$, where $s$ is the split seed. A random ordering of the 871
training targets supplies nested prefixes of 87, 218, 436, 653 and 871
targets. Each prefix refits the training mean, PCA and readouts.
The 218 test targets are unchanged within a split. There are 25
split/repeat runs per fraction. Repeats at 100\% contain the same
training targets and are not treated as additional independent data.

\subsection{Scoring and uncertainty}
\label{app:statistics}

Direction scoring sorts genes by descending absolute observed response,
with a lexical secondary key for ties. RPE1 and Norman use gene names.
K562 uses stringified response-column positions. The same ordering is
used for every predictor within a target. A nonzero prediction receives
one for an observed matching sign and zero otherwise. A zero prediction
receives one half, including when the observed response is zero.
The top-2\% rule is $\max(20,\operatorname{round}(0.02G))$.
The gene set is selected at scoring time, not during fitting or tuning.

For each method and score, prediction and training-mean values are
paired by target and split. Let $\mathcal{S}_t$ be the evaluated splits
containing target $t$, and let $b_t$ be its multiplicity in a bootstrap
sample of unique targets. A bootstrap draw computes
\begin{equation}
\widehat{\Delta}^{\,*}
=\frac{\sum_t b_t\sum_{s\in\mathcal{S}_t}d_{t,s}}
       {\sum_t b_t|\mathcal{S}_t|}.
\label{eq:cluster_bootstrap}
\end{equation}
The original point estimate similarly averages target-split records.
Targets with more evaluated appearances contribute more records, while
all appearances are retained together when resampling. This differs
from the equal-target averaging used for the descriptive cumulative
distributions in Figure~\ref{fig:replication}b.

Percentile intervals use the 2.5th and 97.5th bootstrap percentiles.
Pooled and cross-dataset summaries use 10,000 draws with seed~20260915.
Individual-split and coverage summaries use 2,000 draws with seed~1.
For coverage, the five repeats are averaged within target and split
before applying \Eqref{eq:cluster_bootstrap}.
The intervals quantify target sampling variation conditional on the
evaluated splits and fitted models. They do not resample cells, controls
or the full training procedure.

For direct dataset contrasts, datasets are independently resampled
with seeds~20260915 and~20260916, and the bootstrap effects are
subtracted. Reported contrast intervals are percentile intervals of
those differences. The archived contrast point is the bootstrap mean of the differences and
agrees with the point-estimate difference to within 0.01 percentage
points.

The return package also stores the fraction of bootstrap effects at
or below zero. These are bootstrap-tail summaries, not permutation
test probabilities. The primary Holm family contains six comparisons,
two methods by three gene selections, separately for the all-six and
five-additional-split analyses. Cross-dataset Holm families contain
two methods within each dataset and selection. Direct-contrast Holm
families contain four comparisons per selection. Supportive
Benjamini--Hochberg families contain six comparisons per split scope,
and supportive Bonferroni adjustment uses all 12 rows across the two
scopes. We report paired effects and unadjusted 95\% intervals,
without interpreting those stored tail summaries as null-calibrated
significance tests.

\subsection{Independent-replicate calibration}
\label{app:k562}

The K562 calibration uses all 490 cohort targets and is separate from
the held-out-target prediction experiments. Prediction partitions
0--2 are averaged to obtain a replicate reference $q_t$. Their mean
across cohort targets provides the prediction-side cohort mean
$\hat m$. Mixtures $(1-a)\hat m+aq_t$ are scored against partitions
3 and 4 (modeled in \Eqref{eq:estimated_mean}), then averaged
over targets and evaluation partitions.
This cohort mean is a calibration device, not a training mean
from a particular target holdout. Five gene permutations per target
are reused across its two evaluation partitions.

Figure~\ref{fig:calibration}b displays endpoint-normalized changes
$100[S(a)-S(0)]/[S(1)-S(0)]$ separately for each score.
For panel (c), the displayed sensitivity is
$|S(q,y)-S(\pi q,y)|/|S(q,y)-S(\hat m,y)|$, computed from
cohort-averaged scores. It is a descriptive calibration ratio.
Pearson and Spearman compare gene-aligned response vectors.
Top-20 recall is the overlap of the two sets of 20 largest absolute
responses divided by 20. Rank mean is the gene-count-normalized
average predicted absolute-response rank of the observed top-20 genes,
with average ranks for ties.
The marginal energy and Wasserstein scores discard gene identity
as described in Appendix~\ref{app:proofs}.
Table~\ref{tab:calibration} gives the unnormalized mixture results.

\subsection{Software and reproducibility}

The robustness return records Python 3.10.12, NumPy 1.23.5,
pandas 2.0.3, SciPy 1.10.1, scikit-learn 1.2.2, anndata 0.9.2,
PyTorch \texttt{2.3.0a0+6ddf5cf85e.nv24.04} and matplotlib 3.7.2.
The locked configuration, input manifest, per-target scores,
hyperparameter receipts and scripts identify each reported evaluation.
All tables below are generated directly from the returned CSV files.
The figure-source manifest records those inputs and separates
schematic elements from empirical quantities.

\clearpage
\section{Supporting Results}
\label{app:fullresults}

Tables~\ref{tab:primary_full}--\ref{tab:all_norman} report the primary
RPE1 analyses and the complete coordinated predictor pool.
All effects are prediction minus training-mean direction agreement,
in percentage points. Intervals use target-cluster resampling.
Table~\ref{tab:coverage_full} reports the fixed-test coverage curve,
and Table~\ref{tab:calibration} gives the independent-replicate
calibration in original score units.
\begin{table}[htbp]
\centering
\small
\caption{RPE1 primary paired effects in percentage points with 95\% target-cluster bootstrap intervals. The all-six analysis contains 1,308 records from 792 targets. The five-additional analysis contains 1,090 records from 724 targets. Positive entries indicate a gain over the training mean, and the same convention applies to the subsequent result tables.}
\label{tab:primary_full}
\begin{tabular}{llrr}
\toprule
Split scope & Selection & Kernel ridge & Ridge \\
\midrule
All six & Top 100 & $2.31\;[1.45,3.18]$ & $1.60\;[0.76,2.44]$ \\
All six & Top 2\% & $2.45\;[1.53,3.38]$ & $1.73\;[0.82,2.63]$ \\
All six & Top 20 & $3.57\;[2.44,4.69]$ & $2.86\;[1.72,3.99]$ \\
Five additional & Top 100 & $2.36\;[1.49,3.26]$ & $1.56\;[0.71,2.42]$ \\
Five additional & Top 2\% & $2.50\;[1.55,3.46]$ & $1.68\;[0.76,2.60]$ \\
Five additional & Top 20 & $3.64\;[2.48,4.81]$ & $2.83\;[1.68,3.98]$ \\
\bottomrule
\end{tabular}
\end{table}

\begin{table}[htbp]
\centering
\small
\caption{RPE1 complete direction-score results. Absolute scores are percentages. Paired effects are percentage points with 95\% target-cluster intervals. The training mean is the reference for every paired comparison.}
\label{tab:all_rpe1}
\begin{tabular}{llrr}
\toprule
Predictor & Selection & Score (\%) & Paired effect (pp) \\
\midrule
Training mean & Top 100 & 79.43 & Reference \\
Training mean & Top 2\% & 79.10 & Reference \\
Training mean & Top 20 & 78.94 & Reference \\
Zero & Top 100 & 50.00 & $-29.43\;[-30.87,-27.96]$ \\
Zero & Top 2\% & 50.00 & $-29.10\;[-30.62,-27.57]$ \\
Zero & Top 20 & 50.00 & $-28.94\;[-30.63,-27.27]$ \\
Nearest embedding & Top 100 & 76.61 & $-2.82\;[-4.55,-1.09]$ \\
Nearest embedding & Top 2\% & 76.48 & $-2.62\;[-4.46,-0.78]$ \\
Nearest embedding & Top 20 & 77.58 & $-1.36\;[-3.41,0.66]$ \\
Ridge & Top 100 & 81.02 & $1.60\;[0.76,2.44]$ \\
Ridge & Top 2\% & 80.83 & $1.73\;[0.82,2.63]$ \\
Ridge & Top 20 & 81.80 & $2.86\;[1.72,3.99]$ \\
Kernel ridge & Top 100 & 81.73 & $2.31\;[1.45,3.18]$ \\
Kernel ridge & Top 2\% & 81.54 & $2.45\;[1.53,3.38]$ \\
Kernel ridge & Top 20 & 82.52 & $3.57\;[2.44,4.69]$ \\
Response network & Top 100 & 79.42 & $-0.00\;[-0.01,0.00]$ \\
Response network & Top 2\% & 79.10 & $0.00\;[-0.00,0.00]$ \\
Response network & Top 20 & 78.94 & $0.00\;[0.00,0.01]$ \\
\bottomrule
\end{tabular}
\end{table}

\clearpage

\begin{table}[htbp]
\centering
\small
\caption{K562 complete direction-score results. Absolute scores are percentages. Paired effects are percentage points with 95\% target-cluster intervals. The training mean is the reference for every paired comparison.}
\label{tab:all_k562}
\begin{tabular}{llrr}
\toprule
Predictor & Selection & Score (\%) & Paired effect (pp) \\
\midrule
Training mean & Top 100 & 64.42 & Reference \\
Training mean & Top 2\% & 65.82 & Reference \\
Training mean & Top 20 & 65.82 & Reference \\
Zero & Top 100 & 50.00 & $-14.42\;[-16.21,-12.61]$ \\
Zero & Top 2\% & 50.00 & $-15.82\;[-17.76,-13.85]$ \\
Zero & Top 20 & 50.00 & $-15.82\;[-17.76,-13.85]$ \\
Nearest embedding & Top 100 & 52.40 & $-12.02\;[-13.84,-10.22]$ \\
Nearest embedding & Top 2\% & 53.05 & $-12.77\;[-14.96,-10.55]$ \\
Nearest embedding & Top 20 & 53.05 & $-12.77\;[-14.96,-10.55]$ \\
Ridge & Top 100 & 64.39 & $-0.02\;[-0.15,0.11]$ \\
Ridge & Top 2\% & 65.88 & $0.06\;[-0.19,0.32]$ \\
Ridge & Top 20 & 65.88 & $0.06\;[-0.19,0.32]$ \\
Kernel ridge & Top 100 & 64.29 & $-0.12\;[-0.36,0.11]$ \\
Kernel ridge & Top 2\% & 65.66 & $-0.16\;[-0.58,0.27]$ \\
Kernel ridge & Top 20 & 65.66 & $-0.16\;[-0.58,0.27]$ \\
Response network & Top 100 & 64.38 & $-0.04\;[-0.11,0.03]$ \\
Response network & Top 2\% & 65.77 & $-0.04\;[-0.18,0.09]$ \\
Response network & Top 20 & 65.77 & $-0.04\;[-0.18,0.09]$ \\
\bottomrule
\end{tabular}
\end{table}

\begin{table}[htbp]
\centering
\small
\caption{Norman complete direction-score results. Absolute scores are percentages. Paired effects are percentage points with 95\% target-cluster intervals. The training mean is the reference for every paired comparison.}
\label{tab:all_norman}
\begin{tabular}{llrr}
\toprule
Predictor & Selection & Score (\%) & Paired effect (pp) \\
\midrule
Training mean & Top 100 & 76.95 & Reference \\
Training mean & Top 2\% & 77.12 & Reference \\
Training mean & Top 20 & 75.79 & Reference \\
Zero & Top 100 & 50.00 & $-26.95\;[-32.34,-21.12]$ \\
Zero & Top 2\% & 50.00 & $-27.12\;[-32.48,-21.32]$ \\
Zero & Top 20 & 50.00 & $-25.79\;[-31.90,-19.60]$ \\
Nearest embedding & Top 100 & 71.10 & $-5.86\;[-11.08,-0.61]$ \\
Nearest embedding & Top 2\% & 71.19 & $-5.92\;[-11.15,-0.70]$ \\
Nearest embedding & Top 20 & 68.81 & $-6.98\;[-13.28,-0.71]$ \\
Ridge & Top 100 & 77.89 & $0.94\;[-0.67,2.71]$ \\
Ridge & Top 2\% & 78.00 & $0.88\;[-0.74,2.66]$ \\
Ridge & Top 20 & 77.14 & $1.35\;[-0.83,3.80]$ \\
Kernel ridge & Top 100 & 78.40 & $1.44\;[-1.33,4.43]$ \\
Kernel ridge & Top 2\% & 78.50 & $1.38\;[-1.38,4.36]$ \\
Kernel ridge & Top 20 & 77.38 & $1.59\;[-2.21,5.55]$ \\
Response network & Top 100 & 77.27 & $0.32\;[-0.14,0.78]$ \\
Response network & Top 2\% & 77.42 & $0.30\;[-0.15,0.76]$ \\
Response network & Top 20 & 76.27 & $0.48\;[-0.47,1.39]$ \\
\bottomrule
\end{tabular}
\end{table}

\clearpage

\begin{table}[htbp]
\centering
\small
\caption{Fixed-test RPE1 top-100 effects in percentage points with 95\% target-cluster bootstrap intervals. Five subset repeats are averaged within target and split before inference. Each fraction includes 724 unique test targets across five splits.}
\label{tab:coverage_full}
\begin{tabular}{rrrr}
\toprule
Training (\%) & Targets & Kernel ridge & Ridge \\
\midrule
10 & 87 & $0.01\;[-0.52,0.54]$ & $0.19\;[-0.26,0.64]$ \\
25 & 218 & $1.33\;[0.73,1.93]$ & $1.04\;[0.48,1.59]$ \\
50 & 436 & $1.72\;[0.97,2.45]$ & $1.26\;[0.53,1.97]$ \\
75 & 653 & $2.08\;[1.25,2.90]$ & $1.40\;[0.60,2.18]$ \\
100 & 871 & $2.38\;[1.50,3.25]$ & $1.56\;[0.70,2.39]$ \\
\bottomrule
\end{tabular}
\end{table}

\begin{table}[htbp]
\centering
\small
\caption{Independent-replicate K562 calibration on all 490 targets. Entries are descriptive means over targets and evaluation partitions. The mixture weight controls the noisy reference contribution.}
\label{tab:calibration}
\begin{tabular}{rrrr}
\toprule
Weight & MSE & Pearson & Top-20 recall \\
\midrule
0 & 0.009975 & -0.0257 & 0.0770 \\
0.02 & 0.009942 & -0.0146 & 0.0776 \\
0.05 & 0.009898 & 0.0011 & 0.0891 \\
0.1 & 0.009839 & 0.0232 & 0.1048 \\
0.2 & 0.009777 & 0.0522 & 0.1216 \\
0.35 & 0.009825 & 0.0737 & 0.1303 \\
0.5 & 0.010040 & 0.0837 & 0.1352 \\
0.75 & 0.010770 & 0.0916 & 0.1353 \\
1 & 0.011966 & 0.0953 & 0.1364 \\
\bottomrule
\end{tabular}
\end{table}

\subsection{Ablation of the Auxiliary Penalty}
\label{app:ablation}

Table~\ref{tab:ablation} reports a three-arm ablation of the response
network's training configuration on the six RPE1 splits (three seeds per
arm): the frozen configuration used throughout this paper, the same
configuration with the auxiliary response-to-embedding distance-ratio
penalty removed, and a capacity arm without early stopping. Removing the
penalty restores a response-aligned component in the residual (median
residual--truth correlation $+0.202$, positive in all six splits), with
a smaller amplitude ratio than the frozen configuration. The paired
effect between the two arms, $+0.00008\;[-0.00001,0.00017]$, does not
reach the change produced by a single additional correct gene in one
held-out record, so the reported verdict is unchanged. Removing early stopping raises the
amplitude ratio without a detectable effect, so early stopping is not
the cause. The residual amplitudes are consistent with the bound implied
by the spectral normalization of the network (observed about 1--2\%,
bound about 3--6\%).

That bound follows from the architecture rather than from the data. Each
spectrally normalized layer has spectral norm at most one at the optimum
of its penalty (coefficient $0.01$), a soft constraint rather than an
enforced one, so the four-layer composition has Lipschitz constant at
most one. The network receives L2-normalized principal-component
embeddings, two of which differ by at most two, while a truth residual
with unit marginal variance across $G$ genes has norm of order
$\sqrt{G}$: about $63.7$ for the $4{,}063$ RPE1 genes and $31.6$ for the
$1{,}000$ K562 genes. Target-to-target variation of the required size
would therefore need an amplitude of about $3\%$ on RPE1 and $6\%$ on
K562, and the observed ratios in Table~\ref{tab:ablation}, $0.95\%$,
$0.51\%$ and $1.76\%$, are of this order and below the RPE1 value.

\begin{table}[htbp]
\centering
\caption{Three-arm ablation of the response network's training
configuration on the six RPE1 splits (three seeds per arm). Brackets
show 95\% target-cluster bootstrap intervals. Paired effects are
differences of the top-100 direction agreement against the split-level
train-mean constant, on the fraction scale (the other tables report
percentage points). The residual--truth correlation is the median across
targets of the per-target correlation between the residual (prediction
minus constant) and the truth residual. The amplitude ratio is the
per-target ratio of residual standard deviation (network minus constant)
to residual standard deviation (truth minus constant), summarized as the
median across targets. The ablation covers RPE1 only, and its frozen
configuration re-runs the setting of Section~\ref{sec:rpe1} with three
fit seeds rather than reporting the locked single-seed estimates.}
\label{tab:ablation}
\small
\setlength{\tabcolsep}{4pt}
\begin{tabular}{lccc}
\toprule
Configuration & Paired effect {[CI]} & Resid.\ corr.\ {[CI]} & Amp.\ ratio \\
\midrule
Frozen configuration & $+0.00002\;[-0.00006,+0.00010]$ & $-0.021\;[-0.030,-0.004]$ & $0.95\%$ \\
Auxiliary penalty removed & $+0.00010\;[+0.00004,+0.00016]$ & $+0.202\;[+0.165,+0.213]$ & $0.51\%$ \\
Early stopping removed & $+0.00009\;[-0.00004,+0.00022]$ & $-0.026\;[-0.046,-0.015]$ & $1.76\%$ \\
\bottomrule
\end{tabular}
\end{table}

\clearpage
\section{Coordinated Cross-Dataset Comparison}
\label{app:regimes}

Table~\ref{tab:regimes} applies the same score formula and primary
method family to all three datasets under the coordinated gene
selection $k=\max(20,\operatorname{round}(0.02G))$.
RPE1 retains a positive top-2\% effect for both readouts, and a direct
bootstrap contrast gives an RPE1-minus-K562 difference of 2.61
[1.60, 3.65] percentage points for kernel ridge and 1.66 [0.74, 2.61]
for ridge.

\begin{table}[htbp]
\centering
\caption{Coordinated top-2\% direction-agreement gains over the training
mean, in percentage points. Brackets show 95\% target-cluster bootstrap
intervals. The score selects at least 20 genes. Target populations and
preprocessing remain dataset specific (Table~\ref{tab:design}).}
\label{tab:regimes}
\small
\begin{tabular}{lrrr}
\toprule
Dataset & Unique targets & Kernel ridge & Ridge \\
\midrule
RPE1 & 792 & $2.45\;[1.53,3.38]$ & $1.73\;[0.82,2.63]$ \\
K562 & 432 & $-0.16\;[-0.58,0.27]$ & $0.06\;[-0.19,0.32]$ \\
Norman & 49 & $1.38\;[-1.38,4.36]$ & $0.88\;[-0.74,2.66]$ \\
\bottomrule
\end{tabular}
\end{table}

This comparison identifies a measurable regime difference under a
shared reference and scoring formula.
RPE1 combines positive paired gains with a training-coverage response,
while the K562 estimates place these readouts near the training mean.
Norman extends the comparison to CRISPRa with 49 unique evaluation
targets and a wider interval that includes zero, so the CRISPRa regime
is reported as unresolved rather than as negative evidence.
The within-RPE1 subsampling experiment provides the direct evidence
for training-size sensitivity. Together, these analyses identify the
conditions under which a measured directional gain is reproducible,
which is the model-selection information the benchmark is designed
to supply.

\end{document}